\documentclass{bmvc2k}

\usepackage{times}
\usepackage{epsfig}
\usepackage{graphicx}
\usepackage{amsmath}
\usepackage{amssymb}
\usepackage{multirow}
\usepackage{makecell}
\usepackage{fancyhdr}
\usepackage{arydshln}
\usepackage{animate}
\usepackage{array}
\usepackage{xr}
\usepackage{xparse}

\title{X-Distill: Improving Self-Supervised Monocular Depth via Cross-Task Distillation}

\addauthor{Hong Cai}{hongcai@qti.qualcomm.com}{1}
\addauthor{Janarbek Matai}{jmatai@qti.qualcomm.com}{1}
\addauthor{Shubhankar Borse}{sborse@qti.qualcomm.com}{1}
\addauthor{Yizhe Zhang}{yizhe.zhang.cs@gmail.com}{2}
\addauthor{Amin Ansari}{amina@qti.qualcomm.com}{3}
\addauthor{Fatih Porikli}{fporikli@qti.qualcomm.com}{1}

\addinstitution{
 Qualcomm AI Research\\
 San Diego, CA, USA\\
 Qualcomm AI Research is an initiative of Qualcomm Technologies, Inc.\\
 \textit{Second and third authors contributed equally to this work}
}

\addinstitution{
 Nanjing University of Science and Technology\\
 Nanjing, China\\
 \textit{Work done at Qualcomm AI Research}
}

\addinstitution{
 Qualcomm Technologies, Inc.\\
 San Diego, CA, USA
}

\runninghead{Cai et al.}{X-Distill}

\newcommand{\rev}{}

\begin{document}

\maketitle

\vspace{-10pt}
\begin{abstract}
In this paper, we propose a novel method, X-Distill, to improve the self-supervised training of monocular depth via cross-task knowledge distillation from semantic segmentation to depth estimation. More specifically, during training, we utilize a pretrained semantic segmentation teacher network and transfer its semantic knowledge to the depth network. In order to enable such knowledge distillation across two different visual tasks, we introduce a small, trainable network that translates the predicted depth map to a semantic segmentation map, which can then be supervised by the teacher network. In this way, this small network enables the backpropagation from the semantic segmentation teacher's supervision to the depth network during training. In addition, since the commonly used object classes in semantic segmentation are not directly transferable to depth, we study the visual and geometric characteristics of the objects and design a new way of grouping them that can be shared by both tasks. It is noteworthy that our approach only modifies the training process and does not incur additional computation during inference. We extensively evaluate the efficacy of our proposed approach on the standard KITTI benchmark and compare it with the latest state of the art. We further test the generalizability of our approach on Make3D. Overall, the results show that our approach significantly improves the depth estimation accuracy and outperforms the state of the art. 
\end{abstract}

\vspace{-15pt}
\section{Introduction}\label{sec:intro}
\vspace{-8pt}
Accurate monocular depth estimation plays a critical role in 3D visual scene understanding and is of great importance for a variety of application domains, such as self-driving, AR/VR, and robotics. Thanks to the advancement of deep learning algorithms, recent years have seen considerable progress in this area~\cite{ming2021deep}. However, training accurate deep learning models in a supervised manner requires high-quality (e.g., dense and correctly aligned) ground-truth depth maps, which are difficult and costly to obtain.

In order to overcome this challenge, self-supervision has emerged as a new paradigm for training monocular depth estimation models~\cite{zhou2017unsupervised, godard2017unsupervised, godard2019digging}. 
Since the inception of such self-supervised training, researchers have looked at various directions in order to further improve the depth estimation accuracy, such as designing more complex architectures~\cite{guizilini20203d, johnston2020self, lyu2020hr}, improving the photometric matching~\cite{shu2020feature, jiang2020dipe}, handling dynamic objects~\cite{casser2019depth, gordon2019depth, dai2020self, klingner2020self}, utilizing edge information~\cite{ramirez2018geometry, zhu2020edge, saeedan2021boosting}, multi-task learning~\cite{yin2018geonet, chen2019self, ranjan2019competitive, luo2019every, tosi2020distilled}, and exploiting temporal information~\cite{patil2020don}. 

Given the importance of visual scene understanding for depth estimation, researchers have recently started to study how to utilize semantic segmentation information to improve accuracy. In~\cite{yue2020semi, guizilini2020semantically, kumar2021syndistnet}, the authors use pretrained or jointly trained semantic segmentation networks to assist the depth network during both training and test. While such approaches can considerably improve accuracy, they incur significant extra computation during inference as they require running a separate and usually heavy-weight segmentation network. Another route is to incorporate the semantic information into the loss function, which only requires the extra computation of semantic information during training. One possible way is to include semantic segmentation as an auxiliary task, by co-training a semantic network and a depth network that share a set of layers~\cite{tosi2020distilled}. Other papers compare the semantic segmentations on both the warped and actual versions of a frame, and enforce a consistency regularization~\cite{yang2018segstereo, chen2019towards}. However, this requires running the segmentation network in every training iteration, which still incurs considerable overhead. In~\cite{zhu2020edge}, the authors use the segmentation masks to explicitly regularize the edges on the depth map, but their approach requires semantic labels on the same dataset and introduces many additional hyper-parameters.  

In this paper, we propose a novel cross-task knowledge distillation approach, \textbf{X-Distill}, to utilize semantic information to improve self-supervised monocular depth estimation. Given a pretrained semantic segmentation teacher network, our goal is to transfer the semantic knowledge from this teacher network to the depth network during training, in order to enhance the depth network's visual scene understanding capability. Note that our setting is different from the conventional knowledge distillation where the teacher and student networks share the same visual task. In our case, the outputs of the depth network and the semantic segmentation network are not directly comparable. In order to enable such cross-task distillation, we utilize a small neural network to connect segmentation and depth, by generating semantic segmentation based on the predicted depth. The resulting depth-based semantic segmentation is then supervised by the teacher network. The small network is trained together with the depth network and as such, allows backpropagation from the semantic segmentation teacher's supervision to the depth network.

In addition to enabling gradient flow across the two tasks, it is necessary to redesign the semantic classes such that they are compatible with the visual information in the depth map. In particular, the classes commonly used in semantic segmentation are usually too fine-grained for depth. For instance, road and sidewalk are typically treated as two separate classes in semantic segmentation. However, it is not necessary to treat them differently on the depth map since both of them are on the ground surface and have highly similar depth variation patterns in the field of view. As such, we regroup the objects based on their visual and geometric characteristics. This allows the depth network to distill the key depth-relevant semantic information, without introducing unnecessary difficulties to the learning process. 

We next summarize our main contributions as follows:

\hangindent=24pt 
\textbullet \hspace{2pt} We propose a novel method, X-Distill, to exploit semantic information to improve self-supervised monocular depth estimation. X-Distill enables the depth network to distill semantic knowledge in a cross-task manner from a segmentation teacher network during training. At inference time, the depth network then runs in a standalone manner, without requiring extra computation to process/generate semantic information.

\hangindent=24pt 
\textbullet \hspace{2pt} In order to make the semantic segmentation knowledge compatible with the visual information in depth, we regroup the semantic classes based on the visual and geometric characteristics of the objects. This allows the depth network to distill the key semantic knowledge while removing the unnecessary complexities in the learning.

\hangindent=24pt 
\textbullet \hspace{2pt} We evaluate our proposed approach on KITTI and Make3D, and compare it with the state of the art. We further conduct extensive ablation studies on our method. Overall, our proposed approach achieves considerably more accurate depth estimation, e.g., outperforming~\cite{godard2019digging} by $14\%$ on KITTI (in terms of squared relative error). 

\vspace{-10pt}
\section{Related Work}\label{sec:related work}
\vspace{-8pt}
\noindent \textbf{Self-Supervised Monocular Depth Estimation:}
Due to the difficulty of collecting dense, high-quality ground-truth depth maps, researchers have proposed self-supervised training to obtain monocular depth estimation models. Such self-supervision leverages the geometric relationship among neighboring video frames~\cite{zhou2017unsupervised, godard2019digging} or between the left and right cameras in a stereo setting~\cite{godard2017unsupervised}. While these methods provide a new way to train a depth network without labels, factors such as moving objects, occlusion, poor lighting, and low texture can considerably degrade their performance.

\noindent \textbf{Utilizing Semantic Information for Depth Estimation:} Given the high correlation between semantic and depth information, researchers have studied how to incorporate semantic information to improve depth accuracy. One way is to run an additional (sub)network to generate semantic information at inference time, which can be fed to the depth network~\cite{yue2020semi, guizilini2020semantically, kumar2021syndistnet}. While this approach can considerably improve the depth estimation performance, it incurs significantly more computation. Other works include new loss functions during training, either via multi-task training~\cite{tosi2020distilled} or by enforcing segmentation consistency between the warped and real images~\cite{yang2018segstereo, chen2019towards, meng2019signet}. These methods do not require extra semantic computation during test, but require running a semantic network at every training iteration, which still generates a considerable overhead. 

\noindent \textbf{Knowledge Distillation:} 
Knowledge Distillation is usually used to transfer the knowledge from a more complex model to a smaller model, where both of them are designed for the same visual task~\cite{gou2021knowledge}. Few papers have looked at knowledge distillation across two different visual tasks, e.g., classification tasks with non-overlapping classes~\cite{ye2020distilling}, classification and text-to-image synthesis~\cite{yuan2019ckd}, RGB-based depth estimation and depth super resolution~\cite{sun2021learning}. None of the existing works has studied cross-task distillation from semantic segmentation to depth and we show how to enable it in this paper.

\vspace{-12pt}
\section{Proposed Method}\label{sec:method}
\vspace{-8pt}
In this section, we present our novel take, X-Distill, on utilizing semantic segmentation to improve self-supervised monocular depth estimation, through cross-task distillation. In order to transfer the relevant knowledge from a semantic segmentation teacher network to the depth network during training, we use a small network to translate depth to segmentation, thus enabling gradient flow across the two visual tasks. In addition, we redesign the semantic classes to make them compatible with the visual information contained in depth.

\vspace{-10pt}
\subsection{Self-Supervised Monocular Depth Estimation}\label{sec:ssmde}
\vspace{-5pt}
We utilize self-supervision to train a monocular depth estimation model, based on single-view video sequences~\cite{godard2019digging,zhou2017unsupervised}. 

\noindent \textbf{Geometric Modeling:} Consider two neighboring video frames, $I_t$ and $I_s$. Suppose that pixel $p_t\in I_t$ and pixel $p_s \in I_s$ are two different views of the same point of an object, then $p_t$ and $p_s$ are related geometrically as follows:\vspace{-7pt}
\begin{equation}\label{eq:sfm}
d(p_s)h(p_s) = \mathbf{K} (\mathbf{R}_{t\rightarrow s} \mathbf{K}^{-1}d(p_t)h(p_t) + \mathbf{t}_{t\rightarrow s}),\vspace{-7pt}
\end{equation}
\rev{where $h(p) = [h,\, w,\, 1]$ denotes the homogeneous coordinates of a pixel $p$ with $h$ and $w$ being its vertical and horizontal positions on the image, $d(p)$ is the depth at $p$, $\mathbf{K}\in \mathbb{R}^{3\times3}$ is the camera intrinsic matrix, and $\mathbf{T}_{t\rightarrow s} = [\mathbf{R}_{t\rightarrow s} | \mathbf{t}_{t\rightarrow s}] \in \mathbb{R}^{3\times4}$ is the 6DoF relative camera motion/pose from $t$ to $s$, with $\mathbf{R}_{t\rightarrow s}\in \mathbb{R}^{3\times3}$ and $\mathbf{t}_{t\rightarrow s}\in \mathbb{R}^{3\times1}$ being the rotation matrix and translation vector. }

\rev{Given the depth map} of $I_t$, denoted by $D_t$, and the relative camera pose from $I_t$ to $I_s$, we can synthesize $I_t$ from $I_s$ based on Eq.~\ref{eq:sfm}, assuming that the 3D points captured in $I_t$ are also present in $I_s$. We denote the synthesized/warped version of $I_t$ as $\widehat{I}_t$.

\noindent \textbf{Self-Supervised Training:} Suppose that the depth map and the relative camera pose are provided by a depth network and a pose network, respectively. By minimizing the difference between the warped and actual versions of $I_t$, we can train these two networks. A common photometric loss function for comparing $I_t$ and $\widehat{I}_t$ is given by\vspace{-7pt}
\begin{equation}\label{eq:photometric}
    \mathcal{L}_\text{PH}(I_t,\, \widehat{I}_t) = \alpha \|I_t - \widehat{I}_t\|_1 + (1-\alpha) \frac{1-\text{SSIM}(I_t,\, \widehat{I}_t)}{2},\vspace{-7pt}
\end{equation}
where $\|\cdot\|_1$ denotes the $\mathcal{L}_1$ norm and SSIM is the Structural Similarity Index Measure~\cite{wang2004image}. Note that $\mathcal{L}_\text{PH}$ is computed in a per-pixel manner.

\begin{figure}[t!]
\centering
\includegraphics[width=0.82\linewidth]{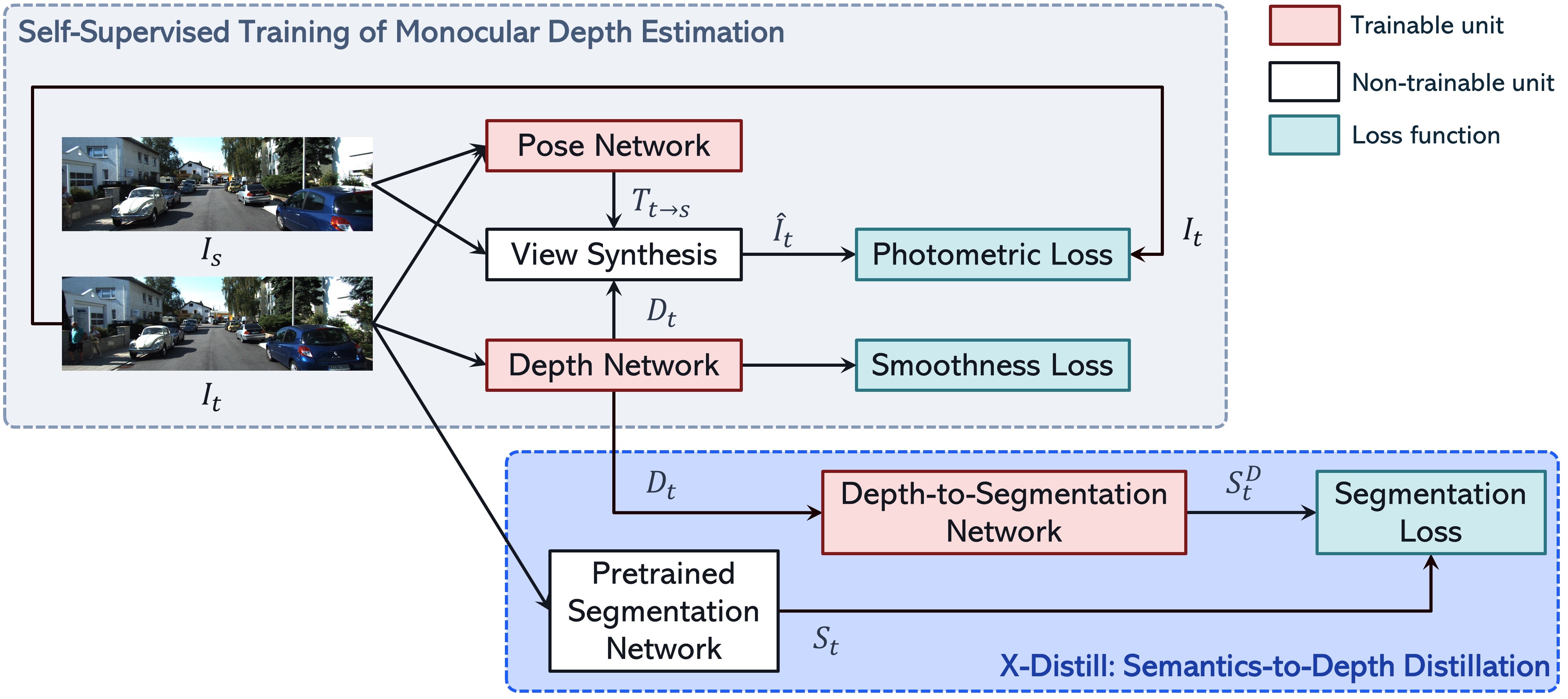}
\vspace{-5pt}
\caption{\small Overview of our proposed X-Distill approach. The gray block describes the self-supervised training of a monocular depth network based on single-view videos. The blue block illustrates our proposed cross-task semantics-to-depth distillation. By utilizing a trainable depth-to-segmentation network to translate predicted depth to segmentation, we enable cross-task knowledge transfer from the pretrained segmentation teacher network to the depth network during training. In addition, we regroup the semantic classes such that they become compatible with the visual information in depth.} 
\label{fig:overview}
\vspace{-12pt}
\end{figure}

It is common to further include a smoothness regularization to prevent drastic variations in the predicted depth map. Furthermore, in practice, not all the 3D points in $I_t$ can be found in $I_s$, due to occlusion and objects (partially) moving out of the frame. Some objects can also be moving (e.g., cars), which is not considered in the geometric model of Eq.~\ref{eq:sfm}. In order to correctly measure the photometric loss and train the networks, it is a common practice to mask out the pixel points that violate the geometric model (see~\cite{godard2019digging} for more details on the masking techniques). Fig.~\ref{fig:overview} (gray block) illustrates the self-supervised training scheme of a monocular depth network.

\vspace{-10pt}
\subsection{Cross-Task Distillation from Semantics to Depth}\label{sec:semantic distillation}
\vspace{-6pt}
Consider a depth network, $f_{D}$, and a pretrained semantic segmentation network, $f_{S}$. Our goal is to transfer the knowledge of the teacher network, $f_{S}$, to the depth network, $f_{D}$. However, unlike conventional knowledge distillation where teacher and student networks are used for the same visual task, $f_{D}$ and $f_{S}$ are used for two different tasks, and their outputs are not directly comparable. In other words, given an input, we cannot directly measure the difference between the outputs of $f_D$ and $f_S$ in order to generate a loss to train $f_D$.

As such, we utilize a Depth-to-Segmentation (D2S) neural network, $h_{D2S}$, to translate depth to semantic segmentation. Given the segmentation map generated from the predicted depth map, we are now able to construct a segmentation loss to distill semantic knowledge from $f_{S}$ to $f_{D}$. More formally, the new loss term is given as follows:\vspace{-7pt}
\begin{equation}\label{eq:d2s loss}
    \mathcal{L}_{D2S}(S_t^\text{D}, S_t) = \sum_{i=1}^{H} \sum_{j=1}^{W} \frac{\mathcal{\mathcal{L}}_\text{CE}(S^\text{D}_t(i,\,j),\,S_t(i,\,j))}{HW},\vspace{-7pt}
\end{equation}
where $S_t^D = h_{D2S}(f_D(I_t))$ is the semantic segmentation map generated by $h_{D2S}$ based on the predicted depth map $D_t = f_D(I_t)$, $S_t$ is the semantic segmentation output generated by the semantic segmentation teacher network, $\mathcal{L}_\text{CE}$ denotes the cross-entropy loss, and $H$ and $W$ are the height and width of the input image.\footnote{\rev{Note that we can include a ``background'' class for the pretrained segmentation model (which is a common practice). This will allow us to ignore pixels that are not of interest when computing the distillation loss of Eq.~\ref{eq:d2s loss}}.}

The total loss is then given by \vspace{-5pt}
\begin{equation}\label{eq:total loss}
    \mathcal{L}_\text{Total} = \sum_{k=1}^{N_s}\mathcal{L}_{PH,\,k} + \sum_{k=1}^{N_s} \lambda_{SM,\,k}\mathcal{L}_{SM,\,k} + \lambda_{D2S}\mathcal{L}_{D2S},\vspace{-5pt}
\end{equation}
where the self-supervised depth loss is computed over $N_s$ scales, $\mathcal{L}_{PH,\,k}$ is the photometric loss at the $k^\text{th}$ scale, $\lambda_{SM,\,k}$ and $\mathcal{L}_{SM,\,k}$ are the weight and loss for the smoothness regularization at the $k^\text{th}$ scale, and $\lambda_{D2S}$ is the weight of the cross-task distillation loss, $\mathcal{L}_{D2S}$.

It can be seen that during training, $h_{D2S}$ is jointly trained with the depth network. This makes it possible for the pretrained teacher network to provide semantic supervision to the depth network, by backpropagating through $h_{D2S}$. Our proposed approach is illustrated in Fig.~\ref{fig:overview}, with the semantics-to-depth distillation module highlighted in the blue block.

For the depth-to-segmentation network, $h_{D2S}$, we adopt a small architecture. More specifically, $h_{D2S}$ consists of two $3\!\times \!3$ convolutional layers, each followed by a BatchNorm layer and a ReLu layer, as well as a pointwise convolutional layer at the end which outputs the segmentation. Note that the $h_{D2S}$ should not be too complex, since a deeper network would take over too much of the learning load and weaken the knowledge flow to the depth network. As we shall see in our experiments in Sec.~\ref{sec:exp}, while using a deeper $h_{D2S}$ can still increase the accuracy of the depth network, the improvement is not as significant as that by using our proposed smaller $h_{D2S}$.  

Once the training is finished, the depth network can then run in a standalone manner, without requiring any extra computation of semantic information during inference. Furthermore, our proposed distillation approach only adds a small amount of computation to training. More specifically, the segmentation maps from the teacher network only need to be computed once and the additional forward/backward passes are cheap since $h_{D2S}$ is small.

\vspace{-5pt}
\subsection{Depth-Compatible Grouping of Semantic Classes}\label{sec:regroup}
\vspace{-5pt}
Semantic segmentation usually contains much more fine-grained visual recognition information that is not present in the depth map. For instance, road and sidewalk are typically treated as two different semantic classes. However, the depth map does not contain such classification information as both road and sidewalk are on the ground plane and have similar depth variations. As a result, it is not necessary to differentiate them on the depth map. On the other hand, the depth map does contain the information for differentiating certain classes. For instance, a road participant (e.g., pedestrian, vehicle) can be easily separated from the background (e.g., road, building) given the different patterns of their depth values.

As such, is it necessary to reconsider the grouping of semantic classes, such that the key semantic information is preserved while the unnecessary complexity is removed from the distillation. Table~\ref{tab:categorization} summarizes our new grouping, which results in four groups.\footnote{We focus on outdoor scenes in this paper and will consider an extension to indoor scenes as part of future work.} In the first two groups, we have objects in the foreground. The respective foreground objects in these two groups are then further differentiated based on their shapes, where the first group contains thin structures, e.g., traffic lights/signs (including the poles), and the second group consists of people and vehicles which are of more general shapes. The third and fourth groups then contain the background objects, such as buildings, vegetation, road, and sidewalk. We further separate the ground plane (e.g., road and sidewalk) from the other background objects.

\begin{table}[t!]
\centering
\footnotesize
\begin{tabular}[h]{ c | c | c | c }
\hline
\multirow{2}{*}{\textbf{Object Groups}} &{\textbf{Foreground}} &{\bf Shape of} &{\bf Location in} \\
& {\bf vs. Background} &{\bf Bounding Box} & {\bf 3D Space} \\
\hline
Thin objects      &Foreground     &Thin rectangular   &Above ground plane\\ \hline
People and vehicles &Foreground     &Rectangular        &Above ground plane\\ \hline
Background objects  &Background     &Not boundable      &Above ground plane\\ \hline
Ground              &Background     &Not boundable      &On ground plane\\
\hline
\end{tabular}
\vspace{2pt}
\caption{\small Depth-compatible semantic class grouping for outdoor scenes.}
\label{tab:categorization}
\vspace{-15pt}
\end{table}

\begin{figure}[b!]
\vspace{-10pt}
\centering
\includegraphics[width=0.95\linewidth]{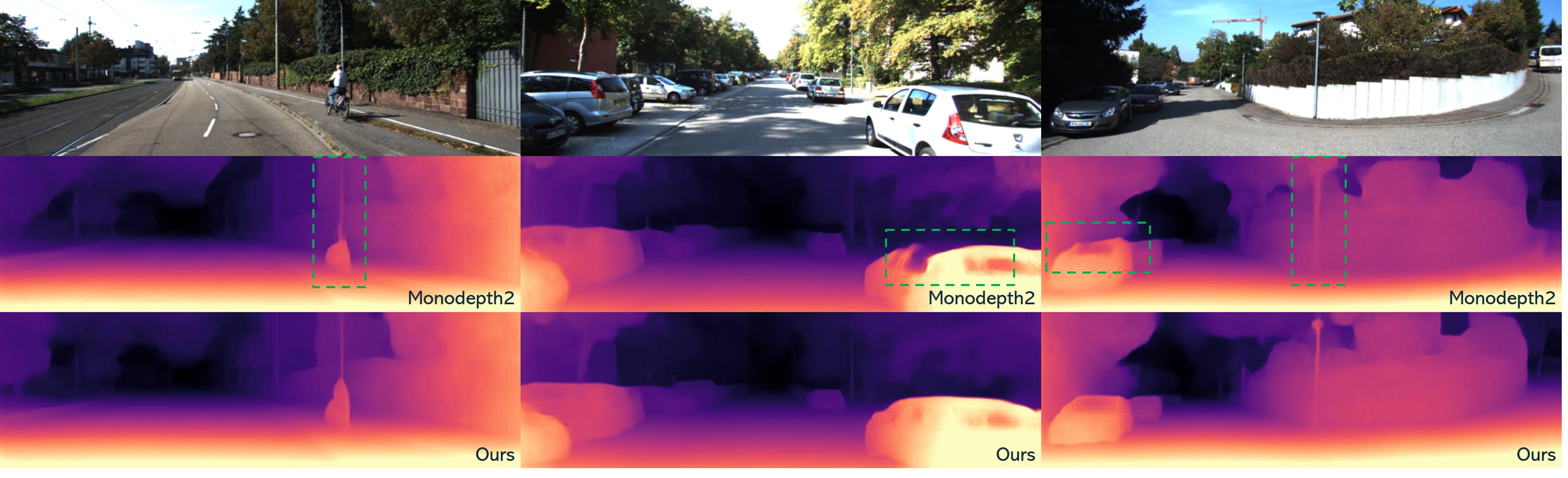}
\vspace{-10pt}
\caption{\small Depth estimation on three sample images. The second row shows the estimated depth maps by Monodepth2~\cite{godard2019digging} and the third row shows the depth maps by our proposed X-Distill approach. It can be seen that our method provides more accurate depth estimation. The green boxes indicate sample regions where our method considerably improves the estimation quality.} \label{fig:sample results}
\vspace{-0pt}
\end{figure} 
 
\vspace{-10pt}
\section{Experiments}\label{sec:exp}
\vspace{-5pt}
In this section (and also in the supplementary file), we present a  comprehensive performance analysis on our proposed X-Distill approach and compare with the current state of the art. We furthermore conduct in-depth ablation studies on various aspects of our method.

\vspace{-10pt}
\subsection{Experiment Setup}
\vspace{-5pt}
\noindent \textbf{Datasets:} We evaluate depth estimation on KITTI~\cite{geiger2013vision} using the standard Eigen split~\cite{eigen2015predicting}, with two input resolutions, $640\!\times \!192$ and $1024\!\times \!320$. Following~\cite{zhou2017unsupervised}, we remove the static frames in the training set. There are 39,810, 4,424, and 697 samples for training, validation, and test.
We use Cityscapes~\cite{cordts2016cityscapes} training and validation sets to train the segmentation teacher network. We further use Make3D~\cite{saxena2005learning, saxena2008make3d} to evaluate the generalizability of our KITTI-trained model.

\noindent \textbf{Grouping of Semantic Classes:} We group the Cityscapes classes according to our proposed scheme in Table~\ref{tab:categorization}, such that they are compatible with the depth information. More specifically, we have 1) thin objects: poles and traffic lights/signs, 2) people and vehicles: persons, riders, cars, trucks, buses, motorcycles, bicycles, and trains, 3) background objects: buildings, walls, fences, vegetation, terrain, and sky, and 4) ground: road and sidewalk. 

\noindent \textbf{Networks:} For the depth network and the pose network, we use the ResNet-50 (RN50)-based models in~\cite{godard2019digging}. The semantic segmentation teacher network is an HRNet~\cite{wang2020deep} with OCR~\cite{yuan2020object} and InverseForm~\cite{borse2021inverseform}. It has an mIoU of 85.6\% on Cityscapes test set. During the self-supervised training of the depth network, this segmentation network is frozen.

\noindent \textbf{Hyperparameters:} For the self-supervised part, we follow the hyperparameter setting in~\cite{godard2019digging}. For the semantics-to-depth distillation loss, $\mathcal{L}_{D2S}$, we linearly increase its weight from 0 to 0.005 during training. As we shall see in the ablation studies, this linear schedule can improve the training as compared to using a constant weight.

\noindent \textbf{Evaluation Metrics:} We use the commonly used error metrics to evaluate the depth estimation performance, including the Absolute Relative Error (Abs Rel), Squared Relative Error (Sq Rel), Root Mean Squared Error (RMSE), and the RMSE of the log of the depth values. In addition, we use the classification metrics, $\delta_1$, $\delta_2$, and $\delta_3$, which measure whether the ratio between the ground-truth and predicted depth values is within a certain interval around 1. Mathematical definitions of these metrics can be found in the supplementary file. 

\vspace{-10pt}
\subsection{Results}
\vspace{-5pt}
We extensively compare our proposed approach with the latest state of the art (SOTA) on KITTI, including methods that 1) use more complex architectures~\cite{guizilini20203d, johnston2020self, lyu2020hr}, 2) require additional computation of semantic information during inference~\cite{guizilini2020semantically}, 3) utilize semantic information during training and do not incur extra computation during test~\cite{klingner2020self}, 4) propose better photometric matching~\cite{jiang2020dipe, shu2020feature}, 5) utilize multiple frames~\cite{patil2020don}, and 6) perform multi-task learning~\cite{tosi2020distilled}. Note that we do not consider pretraining/online finetuning of the depth network or applying post-processing on the predicted depth maps. We also analyze both the depth estimation accuracy and computation efficiency of the methods. Furthermore, we test our KITTI-trained model on Make3D and compare it with the related SOTA to evaluate generalizability. Finally, we perform extensive ablation studies on our proposed approach.

\begin{table}[t!]
\centering
\scriptsize
\begin{tabular}[h]{ c | c | c  c  c  c | c  c  c}
\hline
\multirow{2}{*}{\textbf{Method}} &\multirow{2}{*}{\textbf{Resolution}} &\multicolumn{4}{c|}{\it Lower is Better} &\multicolumn{3}{c}{\it Higher is Better} \\
& &{\bf Abs Rel} &{\bf Sq Rel} &{\bf RMSE} &{\bf $\text{RMSE}_\text{Log}$} &{\bf$\delta_1$} &{\bf$\delta_2$} &{\bf$\delta_3$} 	\\
\hline

Monodepth2~\cite{godard2019digging} (RN18) &$640\times 192$ &0.115 &0.903 &4.863 &0.193 &0.877 &0.959 &0.981 \\

Monodepth2~\cite{godard2019digging} (RN50) &$640\times 192$ &0.110 &0.831 &4.642 &0.187 &0.883 &0.962 &0.982 \\

Tosi et al.~\cite{tosi2020distilled}$^\dagger$ &$640\times 192$ &0.120 &0.792 &4.750 &0.191 &0.856 &0.958 &\textbf{0.984}\\

PackNet-SfM~\cite{guizilini20203d} &$640\times 192$ &0.111 &\underline{0.785} &\underline{4.601} &0.189 &0.878 &0.960 &0.982 \\

Johnston et al.~\cite{johnston2020self} (RN18) &$640\times 192$ &0.111 &0.941 &4.817 &\underline{0.185} &0.885 &0.961 &0.981 \\

Johnston et al.~\cite{johnston2020self} (RN101) &$640\times 192$ &\textbf{0.106} &0.861 &4.699 &\underline{0.185} &\textbf{0.889} &0.962 &0.982 \\

HR-Depth~\cite{lyu2020hr} &$640\times 192$ &\underline{0.109} &0.792 &4.632 &\underline{0.185} &0.884 &0.962 &\underline{0.983} \\

Guizilini et al.~\cite{guizilini2020semantically}$^\dagger$ (RN18) &$640\times 192$ &0.117 &0.854 &4.714 &0.191 &0.877 &0.959 &0.981 \\

Guizilini et al.~\cite{guizilini2020semantically}$^\dagger$ (RN50) &$640\times 192$ &0.113 &0.831 &4.663 &0.189 &0.878 &\textbf{0.971} &0.982 \\

Klingner et al.~\cite{klingner2020self}$^\dagger$ (RN18) &$640\times 192$ &0.113 &0.835 &4.693 &0.191 &0.879 &0.961 &0.981 \\

Klingner et al.~\cite{klingner2020self}$^\dagger$ (RN50) &$640\times 192$ &0.112 &0.833 &4.688 &0.190 &0.884 &0.961 &0.981 \\

DiPE~\cite{jiang2020dipe} &$640\times 192$ &0.112 &0.875 &4.795 &0.190 &0.880 &0.960 &0.981\\

Patil et al.~\cite{patil2020don} &$640\times 192$ &0.111 &0.821 &4.650 &0.187 &0.883 &0.961 &0.982\\

\hdashline
\textbf{X-Distill (ours)}$^\dagger$ &$640\times 192$ &\textbf{0.106} &\textbf{0.777} &\textbf{4.580} &\textbf{0.184} &\underline{0.888} &\underline{0.963} &0.982 \\
\hline

Monodepth2~\cite{godard2019digging} (RN18) &$1024\times 320$ &0.115 &0.882 &4.701 &0.190 &0.879 &0.961 &0.982 \\

Tosi et al.~\cite{tosi2020distilled}$^\dagger$ &$1024\times 320$ &0.118 &0.748 &4.608 &0.186 &0.865 &0.961 &\underline{0.985}\\

PackNet-SfM~\cite{guizilini20203d} &$1280\times 384$ &0.107 &0.802 &4.538 &0.186 &0.886 &0.962 &0.981 \\

HR-Depth~\cite{lyu2020hr} &$1024\times 320$ &0.106 &0.755 &4.472 &0.181 &\underline{0.892} &\textbf{0.966} &0.984 \\

Klingner et al.~\cite{klingner2020self}$^\dagger$ (RN18) &$1280\times 384$ &0.107 &0.768 &\underline{4.468} &0.186 &0.891 &0.963 &0.982 \\

Shu et al.~\cite{shu2020feature} &$1024\times 320$ &\underline{0.104} &\underline{0.729} &4.481 &\textbf{0.179} &0.893 &\underline{0.965} &\textbf{0.987} \\

\hdashline
\textbf{X-Distill (ours)}$^\dagger$ &$1024\times 320$ &\textbf{0.102} &\textbf{0.698} &\textbf{4.439} &\underline{0.180} &\textbf{0.895} &\underline{0.965} &0.983 \\

\hline
\end{tabular}
\vspace{5pt}
\caption{\small Performance evaluation on KITTI Eigen split. For methods that report performance for multiple models, we use the encoder to differentiate them (e.g., RN18 vs. RN50). Note that two architectures can be very different even if they use the same encoder (e.g., Monodepth2~\cite{godard2019digging} vs. Johnston et al.~\cite{johnston2020self}). For each metric, the best (second best) results are in bold (underlined). \rev{We use $^\dagger$ to indicate methods that utilize semantic information during training.}}
\label{tab:results_kitti}
\vspace{-13pt}
\end{table}

\vspace{-10pt}
\subsubsection{Performance Evaluation}
\vspace{-7pt}
\noindent \textbf{Evaluation on KITTI:}
Table~\ref{tab:results_kitti} shows the evaluation results on KITTI and comparison with the latest SOTA methods. It can be seen that our proposed X-Distill approach performs the best for most of the metrics. When our approach does not achieve the top-1 result, it is very close to the best number. Fig.~\ref{fig:sample results} shows sample prediction results of our proposed approach as compared to those by Monodepth2. It can be seen that Monodepth2 can predict inconsistent depth values on an object, which visually appear as missing parts on the depth map (e.g., see the missing upper part of the car in Fig.~\ref{fig:sample results} (middle)). On the other hand, our approach provides more accurate and semantically more structured depth maps, thanks to its ability to better understand the semantics of the scene. For instance, it generates more structurally complete depth estimations for the biker in Fig.~\ref{fig:sample results} (left) and for the cars in the middle and right figures (as indicated by the green boxes). Moreover, our approach is also able to capture the thin structures better. For instance, in Fig.~\ref{fig:sample results} (right), our model is able to generate a more clear depth estimation over the lamp post. This is because the thin objects are grouped as a class in the semantics-to-depth distillation, which encourages the depth network to learn to recognize these structures.

\begin{figure}[h]
\vspace{-14pt}
\centering
\includegraphics[width=0.6\linewidth]{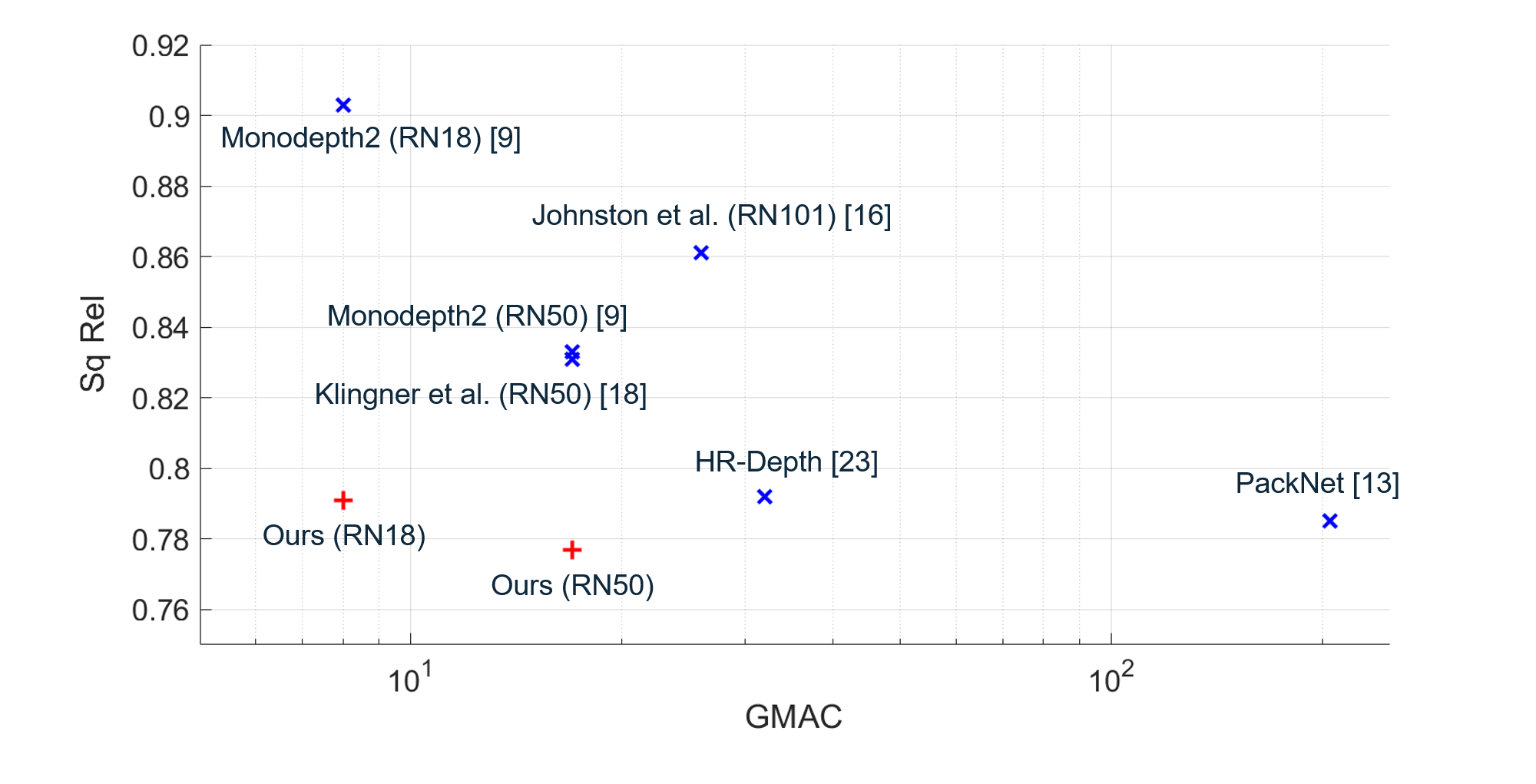}
\vspace{-17pt}
\caption{\small Accuracy (in squared relative error) vs. computation efficiency (in GMAC log-scale).} 
\label{fig:computation}
\vspace{-8pt}
\end{figure}

\vspace{1pt}
\noindent \textbf{Accuracy vs. Computation Efficiency:} Fig.~\ref{fig:computation} shows the accuracy, in terms of squared relative error, and the efficiency, in terms of GMAC (Multiply-Accumulate Operations in $10^9$), of our proposed approach and the SOTA methods on KITTI.\footnote{For Johnston et al.~\cite{johnston2020self}, the GMAC shown in Fig.~\ref{fig:computation} is a lower bound which only includes the RN101 encoder's computation since their self-attention and discrete disparity volume implementation is not publicly available.} It can be seen that our trained model is able to achieve smaller depth estimation errors while using the same or less computation. We further show the performance of applying our cross-task distillation to an RN18-based model from~\cite{godard2019digging}. It can be seen that our method allows this smaller network to achieve an accuracy similar to PackNet (which uses 20$\times$ more computation).

\vspace{1pt}
\noindent \textbf{Depth Estimation on Center and Surrounding Areas:} \rev{Since KITTI images are acquired with a wide-angle lens, we further evaluate the depth estimation performance on center and surrounding areas in the image. Specifically, we horizontally divide each image into 3 equal sections. The middle part is considered the center area and the left and right parts are surrounding areas. It can be seen in Table~\ref{tab:eval_center_vs_side} that the depth estimation is much more accurate in the center area, for both Monodepth2 and X-Distill, since surrounding areas suffer from lens distortion/rectification artifacts. We note that for both areas, our proposed X-Distill consistently provides more accurate depth estimation as compared to Monodepth2.}
\begin{table}[h!]
\vspace{-8pt}
\centering
\scriptsize
\begin{tabular}[h]{ c  | c  c  c  c | c  c  c}
\hline
\multirow{2}{*}{\textbf{Method}} &\multicolumn{4}{c|}{\it Lower is Better} &\multicolumn{3}{c}{\it Higher is Better} \\
&{\bf Abs Rel} &{\bf Sq Rel} &{\bf RMSE} &{\bf $\text{RMSE}_\text{Log}$} &{\bf$\delta_1$} &{\bf$\delta_2$} &{\bf$\delta_3$} \\
\hline
\multicolumn{8}{c}{Over Center Areas}\\ \hline
Monodepth2 (RN50)              &0.061 &0.059 &0.568 &0.077 &0.978 &0.996 &0.999 \\
\textbf{X-Distill (ours)}   &0.056 &0.048 &0.526 &0.072 &0.982 & 0.997 & 0.999 \\ \hline
\multicolumn{8}{c}{Over Surrounding Areas}\\ \hline
Monodepth2 (RN50)           &0.135 &1.228 &5.700 &0.220 &0.839 &0.945 &0.974 \\
\textbf{X-Distill (ours)}    &0.125 &1.054 &5.368 &0.210 &0.852 &0.950 &0.976          \\
\hline
\end{tabular}
\vspace{2pt}
\caption{\small Performance evaluation on center and surrounding image areas.}
\label{tab:eval_center_vs_side}
\vspace{-8pt}
\end{table}

\vspace{1pt}
\noindent \textbf{Generalizability on Make3D:}
We evaluate the generalizability of our KITTI-trained model on Make3D (following the test setup in~\cite{godard2019digging}). It can be seen in Table~\ref{tab:results_make3d} that our model significantly outperforms other SOTA self-supervised methods on this dataset.

\begin{table}[h!]
\vspace{-7pt}
\centering
\scriptsize
\begin{tabular}[h]{ c | c | c  c  c  c}
\hline
\multirow{2}{*}{\textbf{Method}} & \multirow{2}{*}{\textbf{Supervision}}  &\multicolumn{4}{c}{\it Lower is Better} \\ &
& {\bf Abs Rel} &{\bf Sq Rel} &{\bf RMSE} &{\bf $\text{RMSE}_\text{Log}$}\\
\hline

Karsch~\cite{karsch2014depth} &GT &0.428 &5.079 &8.389 &0.149\\
Liu~\cite{liu2014discrete} &GT &0.475 &6.562 &10.05 &0.165\\
Laina~\cite{laina2016deeper} &GT &0.204 &1.840 &5.683 &0.084\\
\hline

Monodepth~\cite{godard2017unsupervised} &S &0.544 &10.94 &11.760 &0.193\\
Zhou et al.~\cite{zhou2017unsupervised} &M &0.383 &5.321 &10.470 &0.478\\
DDVO~\cite{wang2018learning} &M &0.387 &4.720 &8.090 &0.204\\
Monodepth2~\cite{godard2019digging} (RN18) &M &0.322 &3.589 &7.417 &0.163 \\

\hdashline
\textbf{X-Distill (ours)}$^\dagger$ &M &0.308 &3.122 &7.015 &0.158 \\
\hline
\end{tabular}
\vspace{5pt}
\caption{\small Performance evaluation on Make3D. GT indicates that the method is trained with ground-truth Make3D depth annotations, S indicates self-supervised training using KITTI stereo data, and M indicates self-supervised training using KITTI single-view videos. \rev{We use $^\dagger$ to indicate methods that utilize semantic information during training.}}
\label{tab:results_make3d}
\vspace{-10pt}
\end{table}

\vspace{-7pt}
\subsubsection{Ablation Studies}
\vspace{-5pt}

\noindent \textbf{Grouping Semantic Classes:} In addition to our proposed grouping shown in Table~\ref{tab:categorization}, we further test the baseline of using the original 19 Cityscapes classes without regrouping, as well as a more aggressive grouping method that only considers foreground and background objects. As shown in Table~\ref{tab:as_grouping}, while the other two grouping baselines can also improve the depth estimation, the improvements are not as large as compared to our proposed method. 

\begin{table}[h!]
\vspace{-8pt}
\centering
\scriptsize
\begin{tabular}[h]{ c  | c  c  c  c | c  c  c}
\hline
\multirow{2}{*}{\textbf{Categorization Scheme}} &\multicolumn{4}{c|}{\it Lower is Better} &\multicolumn{3}{c}{\it Higher is Better} \\
&{\bf Abs Rel} &{\bf Sq Rel} &{\bf RMSE} &{\bf $\text{RMSE}_\text{Log}$} &{\bf$\delta_1$} &{\bf$\delta_2$} &{\bf$\delta_3$} \\
\hline
Monodepth2 (RN50)              &0.110 &0.831 &4.642 &0.187 &0.883 &0.962 &0.982 \\
\hdashline
Fore/Back-ground (2)       &0.108 &0.798 &4.663 &0.187 &0.886 &0.962 &0.982 \\
Proposed scheme (4)    &0.106 &0.777 &4.580 &0.184 &0.888 &0.963 &0.982 \\
Cityscapes classes (19) &0.110 &0.806 &4.619 &0.184 &0.882 &0.963 &0.983 \\
\hline
\end{tabular}
\vspace{3pt}
\caption{\small Performance of different ways of grouping the semantic classes. }
\label{tab:as_grouping}
\vspace{-7pt}
\end{table}

\noindent \textbf{Complexity of Depth-to-Segmentation Network:}
As discussed in Sec.~\ref{sec:semantic distillation}, the D2S network should be of a proper complexity such that it does not take away the learning from the depth network. As shown in Table~\ref{tab:as_d2s}, by using a more complex D2S network (about 2$\!\times$ larger), the depth network gains a smaller improvement. We further test a baseline using a simple D2S network with one-layer pointwise convolution. This baseline does not perform well as the corresponding D2S network is too simple to translate depth to segmentation.

\begin{table}[h!]
\vspace{-0pt}
\centering
\scriptsize
\begin{tabular}[h]{ c | c  c  c  c | c  c  c}
\hline
{\textbf{Depth-to-Segmentation}}  &\multicolumn{4}{c|}{\it Lower is Better} &\multicolumn{3}{c}{\it Higher is Better} \\
{\textbf{Network}} &{\bf Abs Rel} &{\bf Sq Rel} &{\bf RMSE} &{\bf $\text{RMSE}_\text{Log}$} &{\bf$\delta_1$} &{\bf$\delta_2$} &{\bf$\delta_3$} \\
\hline
Monodepth2 (RN50)     &0.110 &0.831 &4.642 &0.187 &0.883 &0.962 &0.982 \\
\hdashline
2 3$\times$3-Conv + 1 Pointwise Conv  &0.106 &0.777 &4.580 &0.184 &0.888 &0.963 &0.982 \\
4 3$\times$3-Conv + 1 Pointwise Conv  &0.108 &0.786 &4.615 &0.185 &0.886 &0.963 &0.982 \\
1 Pointwise Conv  &0.110 &0.840 &4.683 &0.188 &0.885 &0.961 &0.981 \\
\hline
\end{tabular}
\vspace{3pt}
\caption{\small Depth-to-Segmentation network.}
\label{tab:as_d2s}
\vspace{-12pt}
\end{table}

\noindent \textbf{Weighting Segmentation Loss:} In our proposed approach, we adopt a linear weighting schedule to combine the segmentation distillation loss with the self-supervised depth loss. It can be seen in Table~\ref{tab:as_weights}, the linearly scheduled weight allows the depth network to achieve a higher depth estimation accuracy as compared to using a constant weight. We further vary the final weight by $\pm$20\% and the results show that our proposed method is not very sensitive to the exact value of the weight.
\begin{table}[h!]
\vspace{-5pt}
\centering
\scriptsize
\begin{tabular}[h]{ c | c  c  c  c | c  c  c}
\hline
{\textbf{Weighting of}} &\multicolumn{4}{c|}{\it Lower is Better} &\multicolumn{3}{c}{\it Higher is Better} \\
{\textbf{Distillation Loss}} &{\bf Abs Rel} &{\bf Sq Rel} &{\bf RMSE} &{\bf $\text{RMSE}_\text{Log}$} &{\bf$\delta_1$} &{\bf$\delta_2$} &{\bf$\delta_3$} 	\\
\hline
Monodepth2 (RN50)    &0.110 &0.831 &4.642 &0.187 &0.883 &0.962 &0.982 \\
\hdashline
Constant: 0.0050    &0.109 &0.810 &4.637 &0.185 &0.887 &0.963 &0.983 \\
Linear: 0 - 0.0040  &0.107 &0.779 &4.632 &0.185 &0.887 &0.962 &0.982 \\
Linear: 0 - 0.0045  &0.107 &0.751 &4.553 &0.185 &0.884 &0.963 &0.983 \\
Linear: 0 - 0.0050  &0.106 &0.777 &4.580 &0.184 &0.888 &0.963 &0.982 \\
Linear: 0 - 0.0055  &0.108 &0.795 &4.606 &0.183 &0.887 &0.963 &0.983 \\
Linear: 0 - 0.0060  &0.107 &0.775 &4.580 &0.184 &0.888 &0.963 &0.983 \\
\hline
\end{tabular}
\vspace{3pt}
\caption{\small Weighting of segmentation loss.}
\label{tab:as_weights}
\vspace{-5pt}
\end{table}

\vspace{1pt}
\noindent \textbf{Applying X-Distill to Different Architectures:}  \rev{We apply our proposed approach to different depth networks, e.g., Monodepth2~\cite{godard2019digging} with different encoders and HR-Depth~\cite{lyu2020hr}. Specifically, for the encoder of Monodepth2, in addition to RN18 and RN50 that are used in the original paper, we also employ a recent backbone, DONNA, which is optimized for mobile processors via neural architecture search~\cite{moons2021distilling}. This will demonstrate the efficacy of our method for practical mobile use cases. As can be seen in Table~\ref{tab:as_architecture}, our proposed X-Distill considerably improves the depth estimation accuracy for all these different depth networks.}

\begin{table}[h!]
\vspace{-5pt}
\centering
\scriptsize
\begin{tabular}[h]{ c | c  c  c  c | c  c  c}
\hline
\multirow{2}{*}{\textbf{Architectures}}  &\multicolumn{4}{c|}{\it Lower is Better} &\multicolumn{3}{c}{\it Higher is Better} \\
 &{\bf Abs Rel} &{\bf Sq Rel} &{\bf RMSE} &{\bf $\text{RMSE}_\text{Log}$} &{\bf$\delta_1$} &{\bf$\delta_2$} &{\bf$\delta_3$} \\
\hline
Monodepth2~\cite{godard2019digging} (RN18)  &0.115 &0.903 &4.863 &0.193 &0.877 &0.959 &0.981 \\
\textbf{+ X-Distill} &\textbf{0.111} &\textbf{0.791} &\textbf{4.772} &\textbf{0.188} &0.874 &\textbf{0.960} &\textbf{0.983} \\
\hdashline
Monodepth2~\cite{godard2019digging} (RN50)    &0.110 &0.831 &4.642 &0.187 &0.883 &0.962 &0.982 \\
\textbf{+ X-Distill} &\textbf{0.106} &\textbf{0.777} &\textbf{4.580} &\textbf{0.184} &\textbf{0.888} &\textbf{0.963} &0.982 \\
\hdashline
Monodepth2~\cite{godard2019digging} (DONNA)   &0.115 &0.916 &4.827 &0.193 &0.879 &0.960 &0.981 \\
\textbf{+ X-Distill} &\textbf{0.109} &\textbf{0.772} &\textbf{4.678} &\textbf{0.188} &\textbf{0.884} &\textbf{0.962} &\textbf{0.982} \\
\hdashline
HR-Depth~\cite{lyu2020hr}    &0.109 &0.792 &4.632 &0.185 &0.884 &0.962 &0.983 \\
\textbf{+ X-Distill} &\textbf{0.108} &\textbf{0.755} &\textbf{4.579} &\textbf{0.184} &0.884 &\textbf{0.963} &0.983 \\
\hline
\end{tabular}
\vspace{3pt}
\caption{\small Applying our semantics-to-depth distillation to different depth networks. For each model, improved numbers by using X-Distill are highlighted in bold.}
\label{tab:as_architecture}
\vspace{-5pt}
\end{table}

\vspace{-10pt}
\section{Conclusions}\label{sec:conclusions}
\vspace{-7pt}
In this paper, we presented a novel cross-task distillation approach, X-Distill, to improve the self-supervised training of monocular depth by transferring semantic knowledge from a teacher segmentation network to the depth network. In order to enable such cross-task distillation, we utilized a small, trainable network that translates the predicted depth map to a semantic segmentation map, which the semantic teacher network can then supervise. This enables the backpropagation from the semantic teacher's supervision to the depth network during training. We further studied the visual and geometric characteristics of the objects and designed a new way of grouping them that can be shared by both tasks. We evaluated our proposed approach on KITTI and Make3D, and conducted extensive ablation studies. The results show that by training with the proposed cross-task distillation, we can significantly improve the depth estimation accuracy and outperform the state of the art without incurring additional computation during inference. 


\newpage
\bibliography{egbib}
\end{document}